\pdfoutput=1

\documentclass[11pt]{article}

\usepackage[final]{acl}

\usepackage{times}
\usepackage{latexsym}

\usepackage[T1]{fontenc}

\usepackage[utf8]{inputenc}
\usepackage{colortbl,xcolor} 
\usepackage{graphicx}
\usepackage{microtype}

\usepackage{inconsolata}

\usepackage{graphicx}
\usepackage{adjustbox}
\usepackage{booktabs}

\title{Finetuning End-to-End Models for  Estonian Conversational Spoken Language Translation}

\author{Tiia Sildam, Andra Velve, Tanel Alumäe\\
         Department of Software Science \\
Tallinn University of Technology, Estonia}

\begin{document}
\maketitle
\begin{abstract}
This paper investigates the finetuning of end-to-end models for bidirectional Estonian-English and Estonian-Russian conversational speech-to-text translation. Due to the limited availability of speech translation data for Estonian, we created additional training data by web scraping and synthesizing data from speech recognition datasets using machine translation. We evaluated three publicly available end-to-end models: Whisper, OWSM 3.1, and SeamlessM4T. Our results indicate that fine-tuning with synthetic data enhances translation accuracy by a large margin, with SeamlessM4T matching or surpassing cascaded speech translation systems that use state-of-the-art speech recognition and machine translation models.
\end{abstract}

\section{Introduction}

Estonian language, spoken by around one million native speakers, has benefited significantly from the Estonian Language Technology Program in the last decades \cite{rehm2020european}. This initiative has fostered advancements in several key areas, such as automatic speech recognition (ASR) \cite{alumae-etal-2023-automatic} and machine translation (MT) \cite{tattar2022open}. These improvements are largely due to investments in collecting relevant training data and the successful application of large multilingual pretrained models. Another crucial area of language technology is spoken language translation, which is essential for maintaining smaller languages like Estonian in today's digital world. This technology enables native speakers of a small language to access foreign language content more easily and allows for the broader dissemination of native language content. 
However, one of the significant challenges in developing these technologies is the lack of adequate training data for Estonian, particularly in conversational speech. This shortage hampers the ability to further enhance and refine speech translation tools.

In this study, we explore the finetuning of three publicly available  end-to-end models for bidirectional Estonian-English and Estonian-Russian conversational speech translation tasks and  evaluate their accuracy against the cascaded spoken language translation approach. 
Given the scarcity of speech translation datasets containing significant amounts of conversational speech for these translation directions, we explore two methods to generate additional data: synthesizing speech translation training data from ASR training data using machine translation, and scraping  data (e.g., videos with subtitles) from the internet. We evaluate these models and finetuning approaches using automatic metrics (BLEU and BLEURT) on realistic conversational speech evaluation sets.

The main contribution of this paper is demonstrating that leading large publicly available end-to-end multilingual speech translation models can be fine-tuned to excel in translation tasks involving relatively low-resource languages by using synthetic data generated from diverse ASR training data. Another innovative aspect of the paper is showing that OpenAI's Whisper, originally trained only for translating into English, serves as an effective base model that can be finetuned for other speech translation directions. Additionally, we release an evaluation set for Estonian-English-Russian spoken language translation, which includes conversational speech recordings ``from the wild'', complete with manual transcripts and professionally produced translations\footnote{\url{https://github.com/alumae/k6net6lke-benchmark}}. The best-trained speech translation models are publicly available\footnote{Finetuned Whisper: \url{https://huggingface.co/TalTechNLP/whisper-large-v3-et-en-ru.translate}}. An example of an Estonian TV news broadcast with English and Russian subtitles generated by our finetuned Whisper model is available at \url{https://www.youtube.com/watch?v=rZPqauCYfXI}.

\section{Available models}

In this section, an overview of publicly accessible models suitable for speech translation in the targeted translation directions of our study will be provided.

\subsection{Cascaded spoken language translation}

The cascaded speech translation method involves initially using an ASR system to transcribe speech, followed by translating these transcriptions with a text-to-text MT system. Presently, one of the most widely used multilingual ASR model available to the public is OpenAI's Whisper \cite{radford2022whisper}. In our tests, we utilized the most effective \textit{large-v3} model of Whisper to transcribe English and Russian speech. For Estonian, we used the same model, which was finetuned with 1334 hours of Estonian data available publicly from the TalTech Estonian Speech Dataset 1.0\footnote{\url{https://cs.taltech.ee/staff/tanel.alumae/data/est-pub-asr-data/}} \cite{alumae-etal-2023-automatic}. During the development of this paper, the leading publicly accessible text-to-text MT model for translations involving Estonian was Meta's NLLB-200 \cite{nllbteam2022language}. The NLLB model is available in various sizes, with the largest being the mixture-of-experts (MoE) version, which requires 350 GB of storage. For practical reasons, we opted for the largest dense model, which has 3.3 billion parameters. Machine translation to and from Estonian via text is also well supported by several proprietary vendors via API calls, such as Google and DeepL. The NLP research group at Tartu University offers a publicly accessible NMT system \textit{Neurotõlge}\footnote{\url{https://neurotolge.ee/}} that is effective for Estonian MT tasks \cite{tattar2022open}, and it also provides a free web API for batch processing. OpenAI's GPT models are also capable of conducting machine translation through prompting.

\subsection{End-to-end spoken language translation}

\begin{table*}[tb]
\centering
\begin{tabular}{l|c|rr|rr}
\hline
                     &              & \multicolumn{2}{c|}{CoVoST 2} & \multicolumn{2}{c}{FLEURS} \\
\hline                     
Model                     & \#Parameters & est-eng         & eng-est        & est-eng        & eng-est       \\
\hline
Whisper large-v3     & 1.55B        &  15.0             &    N/A          &  18.7            &      N/A       \\
OWSM 3.1 EBF         & 1B           &    ?           &   7.7           &     ?         &      ?       \\
SeamlessM4T-v2 large & 2.3B         &    27.7           &   29.3           &     31.6         &    22.4        \\
\hline
\end{tabular}
\caption{Speech translation BLEU scores of different publicly available models. N/A denotes that the model is not capable of translating in this direction, and question marks denote scores that are not reported.}
\label{tab:e2e-wer}
\end{table*}

Several publicly available multilingual end-to-end spoken language translation models have recently emerged. OpenAI's Whisper model can perform translation to English from all its supported speech recognition languages.  Other translation directions are not supported by this model. The reported BLEU score for Estonian-to-English translation for the \textit{large-v2} version of Whisper is 18.7, measured on the FLEURS dataset \cite{fleurs2022arxiv} and 15.0, measured on the CoVoST 2 \cite{wang2020covost} dataset.  Both of those datasets contain read speech. Whisper uses Transformer encoder-decoder architecture. 

The Open Whisper-style Speech Model (OWSM) \cite{peng2023owsm} reproduces Whisper-style training using a diverse combination
of publicly available datasets and the open-source toolkit ESPnet \cite{watanabe2018espnet}. It supports multilingual automatic speech recognition
(ASR) and any-to-any speech translation (ST). The latest release of the model (3.1 EBF) uses the E-Branchformer \cite{kim2023branchformer} architecture  in the encoder and Transformer in the decoder. The 1 billion parameter ``base'' version of OWSM 3.1 EBF has a reported BLEU score of 7.7 on the English-to-Estonian translation direction, measured on CoVoST 2.

The third publicly available multilingual speech translation model originates from Meta's SeamlessM4T project \cite{barrault2023seamless}. SeamlessM4T translation models are capable of translating both speech and text modalities, and they can produce both text and speech output. Around 100 languages are supported, although speech output is supported for a much smaller subset of languages. While both Whisper and OWSM models are trained end-to-end from scratch, SeamlessM4T uses a more complicated process for training. First, a self-supervised speech encoder model w2v-BERT 2.0 is pretrained, using a corpus of 4.5M hours of unlabeled audio data covering more than 143 languages. This model is then bridged with the NLLB text-to-text translation model, using special adapter layers that map encoded and time-compressed speech features to the same semantic space as text tokens. This composed model is then finetuned for speech-to-text and speech-to-speech translation tasks, using paired text-text, speech-text and speech-speech data scraped from the web and aligned using a dedicated multimodal embedding and alignment model \cite{duquenne2023sonar}. The  \textit{SeamlessM4T-large-v2} reports a BLEU score of 29.3 on English-Estonian and 27.7 on Estonian-English test sets of CoVoST 2. On FLEURS, this model has a BLEU score of 22.4 on English-Estonian and 31.6 on Estonian-English speech-to-text test sets.

The out-of-the-box BLEU scores of the described models on Estonian-English speech translation tasks are reported in Table \ref{tab:e2e-wer}. Although the scores are measured on test sets containing only read speech, the scores suggest that  these models could be finetuned to perform well also on more conversational speech that is known to be more difficult to translate.

Whisper and OWSM models are designed to handle audio recordings of any length due to the integrated speech segmentation in their decoders. These models effectively generate time-stamped, subtitle-like transcripts, marking each decoded word block with start and end times. In the process of long-form decoding, the models work on 30-second segments of speech at a time, shifting the processing window by 30 seconds (or less) to start where the last decoded word block ended after each decoding step. On the other hand, SeamlessM4T models are limited to processing shorter, utterance-like speech segments, and their translation quality drops substantially with longer segments, often only translating the initial part of the segment. To address this, long recordings must be initially divided into shorter, speaker-consistent segments, typically no longer than 20 seconds, using voice activity detection and speaker segmentation technologies.

\section{Methodology}


The main focus of our work is finetuning publicly available speech translation models using additional data. Since there are no conversational speech translation datasets that include Estonian, we experiment with generating additional data on our own using two methods: web scraping and data synthesis. We compare the performance of all three existing speech translation models before and after finetuning with the same data.

Although Whisper is originally trained to perform only multilingual speech recognition and speech  translation to English, it has been shown that it can perform speech translation to other directions with surprisingly high accuracy by changing only the prefix of the decoder. For example, \citet{peng2023prompting} showed that by only modifying the prompt, Whisper can achieve 18.1 BLEU score on the English-German speech translation test set from the  MuST-C corpus \cite{di2019must}. Therefore, we were relatively confident that Whisper can be finetuned for all translation directions that we were interested in.

The design of Whisper's prompt does not support the specification of alternative translation directions. Consequently, we finetuned Whisper using extra speech translation data by employing the ``transcribe'' prompt, where the language specified in the prompt matched the intended target language. At the inference stage, the expected target language was set in the prompt, but the source language remained unspecified to the model.

On all datasets, Whisper was finetuned\footnote{Finetuning code: \url{https://github.com/alumae/pl-whisper-finetuner}} for three epochs over the additional translation datasets. A learning rate schedule with a peak rate of 1e-04 was used, with  500 warmup steps and a linearly decaying schedule towards 0 after the warmup. An effective batch size of 64 was used. Stochastic weight averaging (SWA) \cite{izmailov2018averaging} with a learning rate of 1e-05 was applied during the last epoch. Adam optimizer was used.

The OWSM 3.1 EBF model underwent finetuning over five epochs, utilizing a batch size of 320 and a maximum learning rate of 2.0e-04, accompanied by a warmup phase of 600 steps. A label smoothing technique was employed with a smoothing factor of 0.1. During training, a multitask encoder-decoder/CTC loss method was used (with source language transcript as supervision for the CTC head), setting the CTC loss weight at 0.3. The majority of these hyperparameters were adopted directly from the ESPnet's training recipe for the OWSM 3.1 EBF model without further adjustments.

The SeamlessM4T model was finetuned using a batch size of 48, peak learning rate of 1e-06 with 100 warmup steps. This finetuning setup integrated automatic early stopping that measured the model's loss on heldout training data after every 1000 model updates and stopped training when the  loss didn't improve during the last 10 evaluations. This usually happened during the second epoch.

For Whisper and OWSM, the training data was compiled to segments of maximally 30 seconds in length, which usually involved concatenating the transcripts of several adjacent utterances from the long-form training audio, together with the corresponding audio chunks (including the audio between transcription end and start times). The SeamlessM4T model was finetuned using the original utterances and/or subtitle segments.

All finetuning experiments were conducted using four Nvidia A100 (80GB) GPUs.

\section{Experimental results}

\subsection{Evaluation data}

A dedicated evaluation dataset was compiled for this project, using data from public sources (e.g. YouTube). When collecting evaluation data, we tried to ensure that it contains mostly long conversational speech recordings with different levels of spontaneousness, such as press conferences, TV talkshows, YouTube videos, and broadcast news with many interviews. Length of evaluation datasets for all directions varied between 3 and 4.6h. Evaluation data is described in Table \ref{tab:evaluation_data}.

\begin{table}
  \centering
  \begin{tabular}{lcc}
    \hline
    \textbf{Direction} & \textbf{Duration} & \textbf{\#Files} \\
    \hline
    Estonian to Eng/Rus & 4.15h & 7 \\
    English to Estonian & 3.05h & 5 \\
    Russian to Estonian & 4.51h & 6 \\
    \hline
  \end{tabular}
  \caption{Amount of evaluation data per translaton direction.}
  \label{tab:evaluation_data}  
\end{table}

Estonian evaluation data was manually transcribed. English and Russian data was all retrieved from YouTube and we relied on the manually created captions of the videos (after some manual post-editing). We took extra care to select such videos that have good quality verbatim captions. The translations for the evaluation data were created by professional translators in Estonia, using both audio transcriptions and audio files as source data.

Table \ref{tab:validation_data_wer} lists ASR word error rates (WER) of Whisper-based models on the evaluation data. The model \textit{whisper-large-v3-est} stands for Whisper's \textit{large-v3} model, finetuned using 1334 hours of Estonian ASR training data. 

WERs were calculated using ASR hypotheses from Whisper's long-form decoding mechanism. Due to that, reference sentences are not aligned with hypotheses. WERs were calculated after removing punctuation, lowercasing both hypotheses and references, and aligning words in the hypotheses with references, using minimum WER segmentation (\textit{mwerSegmenter}) \cite{matusov2005evaluating} via the SLTev toolkit \cite{ansari-etal-2021-sltev}.

It must be noted that Whisper is generally very accurate on English and Russian evaluation data. The surprisingly high WER (compared to the results published by \citet{radford2022whisper}) is mostly caused by occasional hallucinations that repeat some segment transcripts many times.

\begin{table}
  \centering
  \begin{tabular}{llr}
    \hline
    \textbf{Language} & \textbf{Model} & \textbf{WER} \\
    \hline
    English & whisper-large-v3 & 24.5\% \\
    Russian & whisper-large-v3 & 21.1\% \\
    Estonian & whisper-large-v3 & 26.6\% \\
    Estonian & whisper-large-v3-est & 9.7\% \\
    \hline
  \end{tabular}
  \caption{Whisper's speech recognition WER on evaluation data.}
  \label{tab:validation_data_wer}
\end{table}

\subsection{Training data}

In order to finetune the end-to-end speech translation models to perform better in translation directons involving Estonian conversational speech, we experimented with collecting additional data from the web, and synthesizing additional data from ASR training data using MT.

There are some publicly available speech translation datasets that include a relatively small amount of Estonian. The dataset with the largest amount of Estonian is CoVoST 2 with 364 hours of Estonian-English data and 3 hours of English-Estonian data. However, CoVoST 2 includes exclusively read speech and short sentences. The VoxPopuli corpus \cite{wang2021voxpopuli} also contains some Estonian speech, originating from the European Parliamant sessions, but only 3 hours of that are transcribed. Due to the small size or out-of-domain nature, we did not use those datasets for finetuning.

\subsubsection{Scraping web data}

\begin{table}[tb]
    \centering
    \begin{tabular}{lcccc}
        \toprule
        \textbf{Source} & \multicolumn{2}{c}{\textbf{est $\rightarrow$}} & \multicolumn{2}{c}{\textbf{$\rightarrow$ est}} \\
        \cmidrule(lr){2-3} \cmidrule(lr){4-5}
        & \textbf{eng} & \textbf{rus} & \textbf{eng} & \textbf{rus} \\
        \midrule
        ETV+ & - & - & - & 182.7 \\
        TED & - & - & 41.2 & - \\
        TV7 & - & - & 16.4 & - \\
        YouTube & 39.6 & 18.2 & - & 433.9 \\
        \midrule
        & 39.6 & 18.2 & 57.6 & 616.7 \\
        \bottomrule
    \end{tabular}
    \caption{Amount of training data in hours per translation direction, derived from subtitled online videos.}
    \label{tab:web_data_overview}
\end{table}

\begin{table*}[tb]
\centering
\begin{tabular}{l|p{4.2cm}p{4.2cm}p{4.2cm}} \toprule
Language      & \textbf{Estonian} & \textbf{English} & \textbf{Russian} \\ \midrule
Sources      &  \textbf{TalTech Estonian Speech Dataset 1.0 }        & \textbf{Gigaspeech} (subset M):\newline Audiobooks: 260h \newline Podcasts: 350h \newline YouTube: 390 h        & \textbf{DW Russian}: 45h \newline \textbf{TEDx talks:} 57h        \\  \midrule
Total &   1334h       &   1000h      &  102h \\ \bottomrule        
\end{tabular}
\caption{Amount of source-language ASR training data, used as input for creating synthetic speech translation data.}
\label{tab:synth_data}
\end{table*}

Given the relatively small number of Estonian speakers, the amount of speech data available on the web for training speech translation models is limited. We aimed to find data featuring long-form conversational speech (rather than individual utterances) since Whisper and OWSM require 30-second speech segments for training to develop models capable of transcribing long-form speech. We avoided sources with machine-generated subtitles.

We identified several good sources: ETV+ (a Russian-language TV channel of Estonian state media), TED talks with Estonian subtitles, TV7 (an international TV channel with Christian background), and various YouTube channels with consistently good subtitles.

Table \ref{tab:web_data_overview} lists the amount of data we found for each translation direction. As can be seen, the sizes vary significantly across the four translation directions we target.

\subsubsection{Synthetic data}

There are two primary methods for generating synthetic data to train speech translation models: (1) using speech synthesis to create source speech data from existing MT training data, and (2) using MT to generate target text data from existing source language ASR training data. We chose the second method because we already had substantial amount of Estonian ASR training data from various conversational sources, and the current Estonian-to-English and Estonian-to-Russian MT systems produce relatively high-quality translations. The main drawback of the first method is the lack of MT training corpora that include transcribed conversational speech, making it challenging to achieve a wide variety of speakers and natural-sounding speech through speech synthesis.

As Estonian source speech data, we used the data available publicly from the TalTech Estonian Speech Dataset 1.0. It contains mostly speech from broadcast sources, with an emphasis on conversation speech, such as interviews and talk shows. In addition, it contains speech recordings from various conferences and seminars, and a relatively small amount of speech from the Estonian Parliament. All the speech data consists of long-form speech and has been manually transcribed and time-aligned with speech at an utterance level.

When searching for training data for English and Russian speech, we found it challenging to locate high-quality, long-form conversational speech data transcribed at the recording level with orthographic annotation, as needed for finetuning Whisper and OWSM models. For English, we used a subset of the Gigaspeech corpus \cite{chen2021gigaspeech}, which includes long-form recordings (audiobooks, podcasts, and YouTube videos) transcribed at the utterance level. However, these utterances are uppercased, and only a limited set of punctuation marks (``.,!?'') are retained. To enhance the suitability of these transcripts as MT source data, we applied true-casing using a custom implementation. This implementation uses spaCy to split utterances into sentences and then uppercases sentence start tokens, proper nouns, and certain special words (such as \textit{I}).

For Russian, we couldn't find any open datasets that contain sufficient amount of transcribed long-form speech data. A popular choice for training Russian ASR models is the Russian Open STT Dataset\footnote{\url{https://github.com/snakers4/open_stt}} which contains over 20\,000 hours of transcribed Russian speech. However, this dataset contains exclusively relatively short utterances. Although most of the data in this dataset originates from long-form speech recordings, it is not possible to reconstruct homogeneous 30-second speech segments with the corresponding transcripts from this data, as the utterance IDs have been randomized. Therefore, we used two online sources as the Russian speech data, both of which come with good quality captions: Russian TEDx talks and the Russian language YouTube channel of the \textit{Deutsche Welle} (DW) news broadcaster \footnote{\url{https://www.youtube.com/dwrussianreporter}}. 

The total amounts of ASR datasets used as input for synthesizing MT-based speech translation data are listed in Table \ref{tab:synth_data}.
For creating synthetic data for speech translation, the transcripts were machine-translated. We used Google Translate for translating Estonian and English language pair directions. Russian and Estonian language pair translations were done with University of Tartu's \textit{Neurotõlge} MT system. Those choices were based on our budget, as well as  on the reference transcript MT evaluation results in Table \ref{tab:results}.

\begin{table*}[tb]
    \centering
    \setlength{\tabcolsep}{4pt}   
    \begin{adjustbox}{width=\textwidth, totalheight=\textheight, keepaspectratio}
        \begin{tabular}{lccrrrrr|cccccc}
            \toprule
            \textbf{Model} & \multicolumn{2}{c}{\textbf{Finetuned}} & \multicolumn{5}{c}{\textbf{BLEU}} & \multicolumn{5}{c}{\textbf{BLEURT}} \\
            \cmidrule(lr){2-3} \cmidrule(lr){4-8} \cmidrule(lr){9-13}
            & \textbf{web} & \textbf{synt.} & \multicolumn{2}{c}{\textbf{est $\rightarrow$}} & \multicolumn{2}{c}{\textbf{$\rightarrow$ est}} & & \multicolumn{2}{c}{\textbf{est $\rightarrow$}} & \multicolumn{2}{c}{\textbf{$\rightarrow$ est}} \\
            \cmidrule(lr){4-5} \cmidrule(lr){6-7} \cmidrule(lr){9-10} \cmidrule(lr){11-12}
            & & & \textbf{eng} & \textbf{rus} & \textbf{eng} & \textbf{rus} & \textbf{avg} & \textbf{eng} & \textbf{rus} & \textbf{eng} & \textbf{rus} & \textbf{avg} \\
            \midrule
            \midrule            
            \multicolumn{5}{l}{\textit{Text-to-text translation using reference transcripts}} \\
            \midrule
            Ref. + NLLB-200 3.3B & - & - & 31.4 & 25.2 & 21.5 & 19.2 & 24.3 & .652 & .665 & .529 & .574 & .605 \\
            Ref. + GPT3.5-turbo & - & - & 36.1 & 28.3 & 21.3 & 23.8 & 27.4 & .696 & .703 & .593 & .665 & .664  \\
            Ref. + GPT4 & - & - & 38.3 & 31.3 & 19.9 & 24.6 & 28.5 & .702 & .721 & .609 & .656 & .672 \\
            Ref. + Google Translate  & - & - & 38.9 & 26.1 & 25.4 & 24.2 & 28.7 & .690 & .686 & .576 & .655 & .652 \\
            Ref. + Neurotõlge & - & - & 34.8 & 29.3 & 24.7 & 23.7 & 28.1 & .656 & .672 & .558 & .619 & .626 \\
            \midrule
            \midrule            
            \multicolumn{5}{l}{\textit{Cascaded speech translation systems}} \\
            \midrule
            Whisper + NLLB-200 3.3B & - & - & 28.8 & 23.1 & 15.4 & 13.2 & 20.1 & .568 & .568 & .439 & .537 & .528 \\
            Whisper + GPT3.5-turbo & - & - & 32.9 & 26.5 & 15.1 & 18.3 & 23.2 & .649 & .656 & .470 & .621 & .599 \\
            Whisper + GPT4 & - & - & 35.1 & 29.8 & 16.3 & 18.3 & 24.9 & .647 & .687 & .507 & .625 & .617 \\
            Whisper + Google Translate  & - & - & 35.2 & 23.8 & 17.4 & 16.1 & 22.9 & .628 & .617 & .481 & .585 & .578 \\
            Whisper + Neurotõlge & - & - & 31.9 & 26.6 & 16.1 & 16.0 & 22.7 & .598 & .612 & .458 & .566 & .559 \\
            \midrule
            \midrule            
            \multicolumn{5}{l}{\textit{Public end-to-end speech translation models}} \\
            \midrule
            Whisper-large-v3 & - & - & 14.9 & - & - & - & - & .451 & - & - & - & - \\
            OWSM 3.1 EBF & - & - & 0.5 & 0.0 & 1.6 & 0.0 & 0.5 & .176 & .153 & .147 & .095 & .143 \\            
            SeamlessM4T v2 (large) & - & - & 13.2 & 16.2 & 6.4 & 13.9 & 12.4 & .348 & .426 & .227 & .448 & .362 \\
            \midrule
            \midrule
            \multicolumn{8}{l}{\textit{Public end-to-end speech translation models} after finetuning} \\
            \midrule
            Whisper-large-v3 & \checkmark & - & 17.9 & 11.7 & 13.1 & 14.3 & 14.2 & .496 & .413 & .433 & .523 & .466 \\
            Whisper-large-v3 & - & \checkmark & 33.2 & 26.1 & 14.5 & 14.8 & 22.2 & .611 & .605 & .363 & .500 & .520 \\
            Whisper-large-v3 & \checkmark & \checkmark & 33.0 & 25.5 & 17.3 & 16.3 & 23.1 & .614 & .603 & .458 & .549 & .560 \\
            OWSM 3.1 EBF & - & \checkmark & 25.8 & 18.7 & 11.9 & 8.5 & 16.2 & .541 & .463 & .377 & .360 & .435 \\
            SeamlessM4T v2 (large) & \checkmark & - & 19.3 & 14.4 & 6.1 & 4.3 & 11.0 & .468 & .488 & .234 & .261 & .363 \\
            SeamlessM4T v2 (large)  & - & \checkmark & 35.4 & 26.8 & 18.8 & 16.4 & 24.4 & .618 & .603 & .482 & .494 & .549 \\
            SeamlessM4T v2 (large)  & \checkmark & \checkmark & 34.7 & 25.9 & 19.1 & 12.9 & 23.1 & .617 & .605 & .470 & .426 & .529 \\
            \bottomrule
        \end{tabular}
    \end{adjustbox}
    \caption{Comparison of baseline scores, cascaded systems, off-the-shelf end-to-end models and finetuned end-to-end models.}
    \label{tab:results}    
\end{table*}

\subsection{Evaluation metrics}

We based our evaluation on two metrics: BLEU and BLEURT \cite{sellam2020bleurt}. BLEURT is a learned metric, trained on subjective human evaluations scores of machine translation references and the corresponding MT candidates. BLEURT outputs scores that usually in the range of 0..1 (with 1 being a perfect match) and is found to be better correlated with human judgements in several languages. We used the multilingual BLEURT-20-D12 model introduced by \citet{pu2021learning}.

BLEU and BLEURT scores are calculated after aligning words in the translation candidates with references, using \textit{mwerSegmenter} via the SLTev toolkit.

\subsection{Results and discussion}

\begin{table*}[tb]
    \centering
    \setlength{\tabcolsep}{4pt}   
    \begin{adjustbox}{width=\textwidth, totalheight=\textheight, keepaspectratio}
        \begin{tabular}{l|cccc|cccc|cccc}        
            \toprule
            \textbf{Model} &  \multicolumn{4}{c}{\cellcolor{cyan}\textbf{Whisper + Google Translate}} & \multicolumn{4}{c}{\cellcolor{yellow}\textbf{Whisper-large-v3 ft.}} & \multicolumn{4}{c}{\cellcolor{pink}\textbf{SeamlessM4T ft.}} \\
            \midrule
             & est-eng & est-rus & eng-est & rus-est & est-eng & est-rus & eng-est & rus-est & est-eng & est-rus & eng-est & rus-est \\
            \midrule
            \cellcolor{cyan}\textbf{Whisper + Google Translate} &  - & - & - & - &  \cellcolor{cyan}&  \cellcolor{yellow}& &  &  &  \cellcolor{pink}& &  \\
            \cellcolor{yellow}\textbf{Whisper-large-v3 (finetuned)} &  \cellcolor{cyan}&  \cellcolor{yellow}& & & - & - &- & - & &  \cellcolor{pink}& & \\
            \cellcolor{pink}\textbf{SeamlessM4T (finetuned)} & &  \cellcolor{pink}& & &  &  \cellcolor{pink}& &  & - & - &- & -\\            
            \bottomrule
        \end{tabular}
    \end{adjustbox}
    \caption{Statistically significant differences between systems, based on BLEU scores: if one of the models is significantly better than the other, the corresponding cell is colored using the corresponding color.}
    \label{tab:stat_significance}
\end{table*}

Evaluation results, together with several baselines, are presented in Table \ref{tab:results}. 

The first section of rows in the table compares the performance of different MT systems on reference transcripts. It can be seen that while there are substantial differences between the proprietary systems among individual translation directions, the average scores in terms of both BLEU and BLEURT are surprisingly similar. The fully open source NLLB-200 model however doesn't reach the accuracy of the top proprietary systems.

The next section compares MT systems, when using automatically generated transcripts as input. For Russian and English, we used the Whisper \textit{large-v3} model, while for Estonian, the finetuned Whisper model was used. All transcripts were generated using a beam size of 5, with speech activity detection activated in order to exclude non-speech segments from input. It can be seen that for Estonian source speech, using ASR instead of references transcripts deteriorates BLEU scores by around 3 points, while for Russian and English, the decrease in accuracy is larger, which is probably tied to the relatively low WER of Whisper on these datasets, as evident from Table \ref{tab:validation_data_wer}.

The third section of rows compares the out-of-the-box performance of three publicly available end-to-end speech translation models. Whisper produced a segmented transcript directly from the long-form speech recordings, while for OWSM and SeamlessM4T, we segmented the speech into single-speaker chunks using pyannote 3.1 \cite{Plaquet23}. Decoding was performed using beam size of 5 for all models.
The BLEU scores of SeamlessM4T demonstrate the complexity of translating automatically segmented conversational speech, compared to read speech consisting of single utterances: compared to the BLEU scores of the same model on CoVoST 2 and FLEURS test data shown in Table \ref{tab:e2e-wer}, the scores on our evaluation data are lower by a large margin. Contrary to the Estonian-English results on CoVoST 2, Whisper outperforms SeamlessM4T on our data, suggesting that Whisper is better suited for processing conversational speech. OWSM 3.1 EBF, which has a BLEU score of 7.7 on English-Estonian CoVoST 2 data, has close to zero scores on our data in all directions.

The last section of the table compares end-to-end speech translation models after finetuning with synthetic and/or web-scraped data. For Estonian-English and Estonian-Russian, finetuning on synthetic dataset outperforms web data by a large margin, which is expected based on the fact that the Estonian ASR comes from similar domains as evaluation data. In general, SeamlessM4T benefits more than Whisper from finetuning on properly segmented ASR data than from subtitles. This can be explained by the fact that subtitle start and end times are not always properly aligned with speech. For SeamlessM4T, which is finetuned on individual subtitle lines and the corresponding speech segments, this causes the training data to be often corrupted. Whisper, on the other hand, is trained on 30-second chunks of speech that fit typically several lines of subtitles, and the proper subtitle timing is not as important.

Apart from a few outliers, the performance of SeamlessM4T and Whisper are similar, especially in terms of BLEURT scores. This confirms our speculation that Whisper can be finetuned to translate into other directions than it was originally trained for. The performance of OWSM 3.1 EBF is however noticeably lower than for other models after finetuning on synthetic data and in order to save compute time we didn't even finetune it on other datasets.

Since the differences between the BLEU scores from applying different models are relatively small, we used the Wilcoxon signed-rank test to assess whether the difference between the scores was statistically significant. We used BLEU scores of individual evaluation files as input to the paired test. Table \ref{tab:stat_significance} compares the difference between three systems: cascaded system involving Whisper and Google Translate and  Whisper and SeamlessM4T end-to-end models, both finetuned using synthetic speech translation data. It can be seen that the best overall performance is achieved by the finetuned SeamlessM4T model, since no other model is significantly better in any of the directons, while it outperforms both the cascaded system and finetuned Whisper in the Estonian-Russian direction.

Although we haven't performed proper human evaluation of the MT outputs, subjective evaluation by the authors suggests that our best Estonian-English and Estonian-Russian models produce translations that are accurate, fluent and therefore usable in many practical situations (see a translated TV news broadcast at \url{https://www.youtube.com/watch?v=rZPqauCYfXI}). For the opposite direction, the translations have a substantially lower quality by subjective evaluation. These findings correlate with BLEURT scores in Table \ref{tab:results}. 

\section{Conclusion}

In this study, we demonstrated the effectiveness of finetuning end-to-end models for Estonian conversational speech translation using synthetic and web-scraped data. Our experiments revealed that synthetic data derived from ASR training corpora significantly enhances model performance, especially for Whisper and SeamlessM4T models. While all three evaluated models benefited from additional training data, SeamlessM4T worked the most consistently in all directions, indicating its robustness in handling conversational speech translation tasks. The best finetuned models are already usable for Estonian-English and Estonian-English directions for real-world speech data.


The future direction of our research is experimenting with simultaneous speech translation where using end-to-end models is crucial.

\bibliography{reffix}

\begin{thebibliography}{22}
\providecommand{\natexlab}[1]{#1}

\bibitem[{Alum{\"a}e et~al.(2023)Alum{\"a}e, Kalda, Bode, and Kaitsa}]{alumae-etal-2023-automatic}
Tanel Alum{\"a}e, Joonas Kalda, K{\"u}lliki Bode, and Martin Kaitsa. 2023.
\newblock \href {https://aclanthology.org/2023.nodalida-1.49} {Automatic closed captioning for {{E}stonian} live broadcasts}.
\newblock In \emph{The 24th Nordic Conference on Computational Linguistics (NoDaLiDa)}, pages 492--499, T{\'o}rshavn, Faroe Islands.

\bibitem[{Ansari et~al.(2021)Ansari, Bojar, Haddow, and Mahmoudi}]{ansari-etal-2021-sltev}
Ebrahim Ansari, Ondrej Bojar, Barry Haddow, and Mohammad Mahmoudi. 2021.
\newblock \href {https://doi.org/10.18653/V1/2021.EACL-DEMOS.9} {{SLTEV:} comprehensive evaluation of spoken language translation}.
\newblock In \emph{Proceedings of the 16th Conference of the European Chapter of the Association for Computational Linguistics: System Demonstrations, {EACL} 2021}, pages 71--79, Online.

\bibitem[{Chen et~al.(2021)Chen, Chai, Wang, Du, Zhang, Weng, Su, Povey, Trmal, Zhang, Jin, Khudanpur, Watanabe, Zhao, Zou, Li, Yao, Wang, You, and Yan}]{chen2021gigaspeech}
Guoguo Chen, Shuzhou Chai, Guan{-}Bo Wang, Jiayu Du, Wei{-}Qiang Zhang, Chao Weng, Dan Su, Daniel Povey, Jan Trmal, Junbo Zhang, Mingjie Jin, Sanjeev Khudanpur, Shinji Watanabe, Shuaijiang Zhao, Wei Zou, Xiangang Li, Xuchen Yao, Yongqing Wang, Zhao You, and Zhiyong Yan. 2021.
\newblock \href {https://doi.org/10.21437/INTERSPEECH.2021-1965} {{GigaSpeech}: An evolving, multi-domain {ASR} corpus with 10\,000 hours of transcribed audio}.
\newblock In \emph{Interspeech 2021, 22nd Annual Conference of the International Speech Communication Association}, pages 3670--3674, Brno, Czechia.

\bibitem[{Conneau et~al.(2022)Conneau, Ma, Khanuja, Zhang, Axelrod, Dalmia, Riesa, Rivera, and Bapna}]{fleurs2022arxiv}
Alexis Conneau, Min Ma, Simran Khanuja, Yu~Zhang, Vera Axelrod, Siddharth Dalmia, Jason Riesa, Clara Rivera, and Ankur Bapna. 2022.
\newblock \href {https://doi.org/10.1109/SLT54892.2023.10023141} {{FLEURS}: Few-shot learning evaluation of universal representations of speech}.
\newblock In \emph{{IEEE} Spoken Language Technology Workshop, {SLT} 2022}, pages 798--805, Doha, Qatar.

\bibitem[{Duquenne et~al.(2023)Duquenne, Schwenk, and Sagot}]{duquenne2023sonar}
Paul{-}Ambroise Duquenne, Holger Schwenk, and Beno{\^{\i}}t Sagot. 2023.
\newblock \href {https://doi.org/10.48550/ARXIV.2308.11466} {{SONAR:} sentence-level multimodal and language-agnostic representations}.
\newblock \emph{CoRR}, abs/2308.11466.

\bibitem[{Gangi et~al.(2019)Gangi, Cattoni, Bentivogli, Negri, and Turchi}]{di2019must}
Mattia Antonino~Di Gangi, Roldano Cattoni, Luisa Bentivogli, Matteo Negri, and Marco Turchi. 2019.
\newblock \href {https://doi.org/10.18653/V1/N19-1202} {{MuST-C}: a multilingual speech translation corpus}.
\newblock In \emph{Proceedings of the 2019 Conference of the North American Chapter of the Association for Computational Linguistics: Human Language Technologies, {NAACL-HLT} 2019, Minneapolis, MN, Volume 1 (Long and Short Papers)}, pages 2012--2017, USA.

\bibitem[{Izmailov et~al.(2018)Izmailov, Podoprikhin, Garipov, Vetrov, and Wilson}]{izmailov2018averaging}
Pavel Izmailov, Dmitrii Podoprikhin, Timur Garipov, Dmitry~P. Vetrov, and Andrew~Gordon Wilson. 2018.
\newblock \href {http://auai.org/uai2018/proceedings/papers/313.pdf} {Averaging weights leads to wider optima and better generalization}.
\newblock In \emph{Proceedings of the Thirty-Fourth Conference on Uncertainty in Artificial Intelligence, {UAI} 2018}, pages 876--885, Monterey, California, USA.

\bibitem[{Kim et~al.(2022)Kim, Wu, Peng, Pan, Sridhar, Han, and Watanabe}]{kim2023branchformer}
Kwangyoun Kim, Felix Wu, Yifan Peng, Jing Pan, Prashant Sridhar, Kyu~Jeong Han, and Shinji Watanabe. 2022.
\newblock \href {https://doi.org/10.1109/SLT54892.2023.10022656} {{E-Branchformer}: Branchformer with enhanced merging for speech recognition}.
\newblock In \emph{{IEEE} Spoken Language Technology Workshop, {SLT} 2022}, pages 84--91, Doha, Qatar.

\bibitem[{Matusov et~al.(2005)Matusov, Leusch, Bender, and Ney}]{matusov2005evaluating}
Evgeny Matusov, Gregor Leusch, Oliver Bender, and Hermann Ney. 2005.
\newblock \href {http://www.isca-speech.org/archive/iwslt\_05/slt5\_138.html} {Evaluating machine translation output with automatic sentence segmentation}.
\newblock In \emph{2005 International Workshop on Spoken Language Translation, {IWSLT} 2005}, pages 138--144, Pittsburgh, PA, USA.

\bibitem[{{NLLB Team} et~al.(2022){NLLB Team}, Costa-jussà, Cross, Çelebi, Elbayad, Heafield, Heffernan, Kalbassi, Lam, Licht, Maillard, Sun, Wang, Wenzek, Youngblood, Akula, Barrault, Gonzalez, Hansanti, Hoffman, Jarrett, Sadagopan, Rowe, Spruit, Tran, Andrews, Ayan, Bhosale, Edunov, Fan, Gao, Goswami, Guzmán, Koehn, Mourachko, Ropers, Saleem, Schwenk, and Wang}]{nllbteam2022language}
{NLLB Team}, Marta~R. Costa-jussà, James Cross, Onur Çelebi, Maha Elbayad, Kenneth Heafield, Kevin Heffernan, Elahe Kalbassi, Janice Lam, Daniel Licht, Jean Maillard, Anna Sun, Skyler Wang, Guillaume Wenzek, Al~Youngblood, Bapi Akula, Loic Barrault, Gabriel~Mejia Gonzalez, Prangthip Hansanti, John Hoffman, Semarley Jarrett, Kaushik~Ram Sadagopan, Dirk Rowe, Shannon Spruit, Chau Tran, Pierre Andrews, Necip~Fazil Ayan, Shruti Bhosale, Sergey Edunov, Angela Fan, Cynthia Gao, Vedanuj Goswami, Francisco Guzmán, Philipp Koehn, Alexandre Mourachko, Christophe Ropers, Safiyyah Saleem, Holger Schwenk, and Jeff Wang. 2022.
\newblock \href {https://arxiv.org/abs/2207.04672} {No language left behind: Scaling human-centered machine translation}.
\newblock \emph{Preprint}, arXiv:2207.04672.

\bibitem[{Peng et~al.(2023{\natexlab{a}})Peng, Yan, Watanabe, and Harwath}]{peng2023prompting}
Puyuan Peng, Brian Yan, Shinji Watanabe, and David Harwath. 2023{\natexlab{a}}.
\newblock \href {https://doi.org/10.48550/ARXIV.2305.11095} {Prompting the hidden talent of web-scale speech models for zero-shot task generalization}.
\newblock \emph{CoRR}, abs/2305.11095.

\bibitem[{Peng et~al.(2023{\natexlab{b}})Peng, Tian, Yan, Berrebbi, Chang, Li, Shi, Arora, Chen, Sharma, Zhang, Sudo, Shakeel, Jung, Maiti, and Watanabe}]{peng2023owsm}
Yifan Peng, Jinchuan Tian, Brian Yan, Dan Berrebbi, Xuankai Chang, Xinjian Li, Jiatong Shi, Siddhant Arora, William Chen, Roshan~S. Sharma, Wangyou Zhang, Yui Sudo, Muhammad Shakeel, Jee{-}Weon Jung, Soumi Maiti, and Shinji Watanabe. 2023{\natexlab{b}}.
\newblock \href {https://doi.org/10.1109/ASRU57964.2023.10389676} {Reproducing {W}hisper-style training using an open-source toolkit and publicly available data}.
\newblock In \emph{{IEEE} Automatic Speech Recognition and Understanding Workshop, {ASRU} 2023}, pages 1--8, Taipei, Taiwan.

\bibitem[{Plaquet and Bredin(2023)}]{Plaquet23}
Alexis Plaquet and Herv{\'{e}} Bredin. 2023.
\newblock \href {https://doi.org/10.48550/ARXIV.2310.13025} {Powerset multi-class cross entropy loss for neural speaker diarization}.
\newblock \emph{CoRR}, abs/2310.13025.

\bibitem[{Pu et~al.(2021)Pu, Chung, Parikh, Gehrmann, and Sellam}]{pu2021learning}
Amy Pu, Hyung~Won Chung, Ankur~P. Parikh, Sebastian Gehrmann, and Thibault Sellam. 2021.
\newblock \href {https://doi.org/10.18653/V1/2021.EMNLP-MAIN.58} {Learning compact metrics for {MT}}.
\newblock In \emph{Proceedings of the 2021 Conference on Empirical Methods in Natural Language Processing, {EMNLP} 2021, Virtual Event /}, pages 751--762, Punta Cana, Dominican Republic.

\bibitem[{Radford et~al.(2023)Radford, Kim, Xu, Brockman, McLeavey, and Sutskever}]{radford2022whisper}
Alec Radford, Jong~Wook Kim, Tao Xu, Greg Brockman, Christine McLeavey, and Ilya Sutskever. 2023.
\newblock \href {https://proceedings.mlr.press/v202/radford23a.html} {Robust speech recognition via large-scale weak supervision}.
\newblock In \emph{International Conference on Machine Learning, {ICML} 2023, 23-29 July 2023}, volume 202 of \emph{Proceedings of Machine Learning Research}, pages 28492--28518, Honolulu, Hawaii, USA.

\bibitem[{Rehm et~al.(2020)Rehm, Marheinecke, Hegele, Piperidis, Bontcheva, Hajic, Choukri, Vasiljevs, Backfried, Prinz, G{\'{o}}mez{-}P{\'{e}}rez, Meertens, Lukowicz, van Genabith, L{\"{o}}sch, Slusallek, Irgens, Gatellier, K{\"{o}}hler, Bars, Anastasiou, Auksoriute, Bel, Branco, Budin, Daelemans, Smedt, Garab{\'{\i}}k, Gavriilidou, Gromann, Koeva, Krek, Krstev, Lind{\'{e}}n, Magnini, Odijk, Ogrodniczuk, R{\"{o}}gnvaldsson, Rosner, Pedersen, Skadina, Tadic, Tufis, V{\'{a}}radi, Vider, Way, and Yvon}]{rehm2020european}
Georg Rehm, Katrin Marheinecke, Stefanie Hegele, Stelios Piperidis, Kalina Bontcheva, Jan Hajic, Khalid Choukri, Andrejs Vasiljevs, Gerhard Backfried, Christoph Prinz, Jos{\'{e}}~Manu{\'{e}}l G{\'{o}}mez{-}P{\'{e}}rez, Luc Meertens, Paul Lukowicz, Josef van Genabith, Andrea L{\"{o}}sch, Philipp Slusallek, Morten Irgens, Patrick Gatellier, Joachim K{\"{o}}hler, Laure~Le Bars, Dimitra Anastasiou, Albina Auksoriute, N{\'{u}}ria Bel, Ant{\'{o}}nio Branco, Gerhard Budin, Walter Daelemans, Koenraad~De Smedt, Radovan Garab{\'{\i}}k, Maria Gavriilidou, Dagmar Gromann, Svetla Koeva, Simon Krek, Cvetana Krstev, Krister Lind{\'{e}}n, Bernardo Magnini, Jan Odijk, Maciej Ogrodniczuk, Eir{\'{\i}}kur R{\"{o}}gnvaldsson, Mike Rosner, Bolette~S. Pedersen, Inguna Skadina, Marko Tadic, Dan Tufis, Tam{\'{a}}s V{\'{a}}radi, Kadri Vider, Andy Way, and Fran{\c{c}}ois Yvon. 2020.
\newblock \href {https://aclanthology.org/2020.lrec-1.407/} {The {European} language technology landscape in 2020: Language-centric and human-centric {AI} for cross-cultural communication in multilingual {E}urope}.
\newblock In \emph{Proceedings of The 12th Language Resources and Evaluation Conference, {LREC} 2020}, pages 3322--3332, Marseille, France.

\bibitem[{{Seamless Communication} et~al.(2023){Seamless Communication}, Barrault, Chung, Meglioli, Dale, Dong, Duppenthaler, Duquenne, Ellis, Elsahar, Haaheim et~al.}]{barrault2023seamless}
{Seamless Communication}, Lo{\"\i}c Barrault, Yu-An Chung, Mariano~Coria Meglioli, David Dale, Ning Dong, Mark Duppenthaler, Paul-Ambroise Duquenne, Brian Ellis, Hady Elsahar, Justin Haaheim, et~al. 2023.
\newblock Seamless: Multilingual expressive and streaming speech translation.
\newblock \emph{arXiv preprint arXiv:2312.05187}.

\bibitem[{Sellam et~al.(2020)Sellam, Das, and Parikh}]{sellam2020bleurt}
Thibault Sellam, Dipanjan Das, and Ankur~P. Parikh. 2020.
\newblock \href {https://doi.org/10.18653/V1/2020.ACL-MAIN.704} {{BLEURT:} learning robust metrics for text generation}.
\newblock In \emph{Proceedings of the 58th Annual Meeting of the Association for Computational Linguistics, {ACL} 2020}, pages 7881--7892, Online.

\bibitem[{T{\"{a}}ttar et~al.(2022)T{\"{a}}ttar, Purason, Kuulmets, Luhtaru, R{\"{a}}tsep, Tars, Pinnis, Bergmanis, and Fishel}]{tattar2022open}
Andre T{\"{a}}ttar, Taido Purason, Hele{-}Andra Kuulmets, Agnes Luhtaru, Liisa R{\"{a}}tsep, Maali Tars, Marcis Pinnis, Toms Bergmanis, and Mark Fishel. 2022.
\newblock \href {https://doi.org/10.22364/BJMC.2022.10.3.15} {Open and competitive multilingual neural machine translation in production}.
\newblock \emph{Balt. J. Mod. Comput.}, 10(3).

\bibitem[{Wang et~al.(2021)Wang, Rivi{\`{e}}re, Lee, Wu, Talnikar, Haziza, Williamson, Pino, and Dupoux}]{wang2021voxpopuli}
Changhan Wang, Morgane Rivi{\`{e}}re, Ann Lee, Anne Wu, Chaitanya Talnikar, Daniel Haziza, Mary Williamson, Juan~Miguel Pino, and Emmanuel Dupoux. 2021.
\newblock \href {https://doi.org/10.18653/V1/2021.ACL-LONG.80} {{VoxPopuli}: {A} large-scale multilingual speech corpus for representation learning, semi-supervised learning and interpretation}.
\newblock In \emph{Proceedings of the 59th Annual Meeting of the Association for Computational Linguistics and the 11th International Joint Conference on Natural Language Processing, {ACL/IJCNLP} 2021, (Volume 1: Long Papers)}, pages 993--1003, Virtual Event.

\bibitem[{Wang et~al.(2020)Wang, Wu, and Pino}]{wang2020covost}
Changhan Wang, Anne Wu, and Juan~Miguel Pino. 2020.
\newblock \href {https://arxiv.org/abs/2007.10310} {{CoVoST} 2: {A} massively multilingual speech-to-text translation corpus}.
\newblock \emph{CoRR}, abs/2007.10310.

\bibitem[{Watanabe et~al.(2018)Watanabe, Hori, Karita, Hayashi, Nishitoba, Unno, Soplin, Heymann, Wiesner, Chen, Renduchintala, and Ochiai}]{watanabe2018espnet}
Shinji Watanabe, Takaaki Hori, Shigeki Karita, Tomoki Hayashi, Jiro Nishitoba, Yuya Unno, Nelson Enrique~Yalta Soplin, Jahn Heymann, Matthew Wiesner, Nanxin Chen, Adithya Renduchintala, and Tsubasa Ochiai. 2018.
\newblock \href {https://doi.org/10.21437/INTERSPEECH.2018-1456} {{ESPnet}: End-to-end speech processing toolkit}.
\newblock In \emph{Interspeech 2018, 19th Annual Conference of the International Speech Communication Association}, pages 2207--2211, Hyderabad, India.

\end{thebibliography}

\end{document}